\pdfoutput=1

\documentclass[11pt]{article}

\usepackage{acl}

\usepackage{times}
\usepackage{latexsym}

\usepackage[T1]{fontenc}

\usepackage[utf8]{inputenc}

\usepackage{microtype}

\usepackage{latexsym} 
\usepackage{graphicx}
\usepackage{wrapfig}
\usepackage{amssymb}
\usepackage{color,xcolor,colortbl}
\usepackage{enumitem}
\usepackage{soul}
\usepackage{amsmath}
\usepackage{url}
\usepackage{algorithm}
\usepackage[noend]{algpseudocode}
\usepackage{booktabs}
\usepackage{bm,bbm}
\usepackage{mathtools}
\usepackage{array}
\usepackage{multirow}
\usepackage{subcaption}
\usepackage{pbox}
\usepackage{pifont}
\usepackage{arydshln}
\usepackage{dblfloatfix}
\usepackage{relsize}
\usepackage{setspace}

\usepackage{pifont}

\newcommand\Tstrut{\rule{0pt}{2.5ex}}     

\newcommand\numberthis{\addtocounter{equation}{1}\tag{\theequation}}

\definecolor{darkblue}{rgb}{0.0, 0.0, 0.55}
\newenvironment{fontpbk}{\fontfamily{lmss}\selectfont}{\par} 
\setulcolor{blue} 
\usepackage[T1]{fontenc}
\usepackage[scaled=.72]{beramono}

\usepackage{soul}

\definecolor{d}{HTML}{c6dbef}
\definecolor{l}{HTML}{e5f5e0}
\definecolor{mygreen}{rgb}{0.13, 0.55, 0.13}

\title{Towards Abstractive Grounded Summarization of Podcast Transcripts}

\author{
Kaiqiang Song,$^{\dag\ddag}$ Chen Li,$^\ddag$ Xiaoyang Wang,$^\ddag$ Dong Yu,$^\ddag$ Fei Liu$^\dag$\\[0.5em]
$^\ddag$Tencent AI Lab, Seattle, WA\\
$^\dag$University of Central Florida, Orlando, FL\\[0.3em]
\texttt{\{riversong,ailabchenli,shawnxywang,dyu\}@tencent.com}\\
\texttt{feiliu@cs.ucf.edu}\\
}

\begin{document}
\maketitle
\begin{abstract}

Podcasts have shown a recent rise in popularity.
Summarization of podcasts is of practical benefit to both content providers and consumers.
It helps people quickly decide whether they will listen to a podcast and/or reduces the cognitive load of content providers to write summaries.
Nevertheless, podcast summarization faces significant challenges including factual inconsistencies of summaries with respect to the inputs. 
The problem is exacerbated by speech disfluencies and recognition errors in transcripts of spoken language.
In this paper, we explore a novel abstractive summarization method to alleviate these issues. 
Our approach learns to produce an abstractive summary while grounding summary segments in specific regions of the transcript to allow for full inspection of summary details.
We conduct a series of analyses of the proposed approach on a large podcast dataset and show that the approach can achieve promising results.
Grounded summaries bring clear benefits in locating the summary and transcript segments that contain inconsistent information, and hence improve summarization quality in terms of automatic and human evaluation.

\end{abstract}

\section{Introduction}

Podcasts are one of the most popular forms of new media.
As of today, over 155 million people listen to a podcast every week~\cite{Christian:2021}.
With the growing interest, there is an increased demand for textual summaries that foretell the content of podcasts.
Those summaries help people decide, in a few seconds, if they will listen to a podcast or subscribe to the channel. 
They are helpful for users who want to find podcasts previously listened to.
Furthermore, they can be re-purposed for social media posts or email marketing campaigns, enabling content creators to make their podcasts accessible to a larger audience.

\begin{figure*}
\centering
\includegraphics[width=6in]{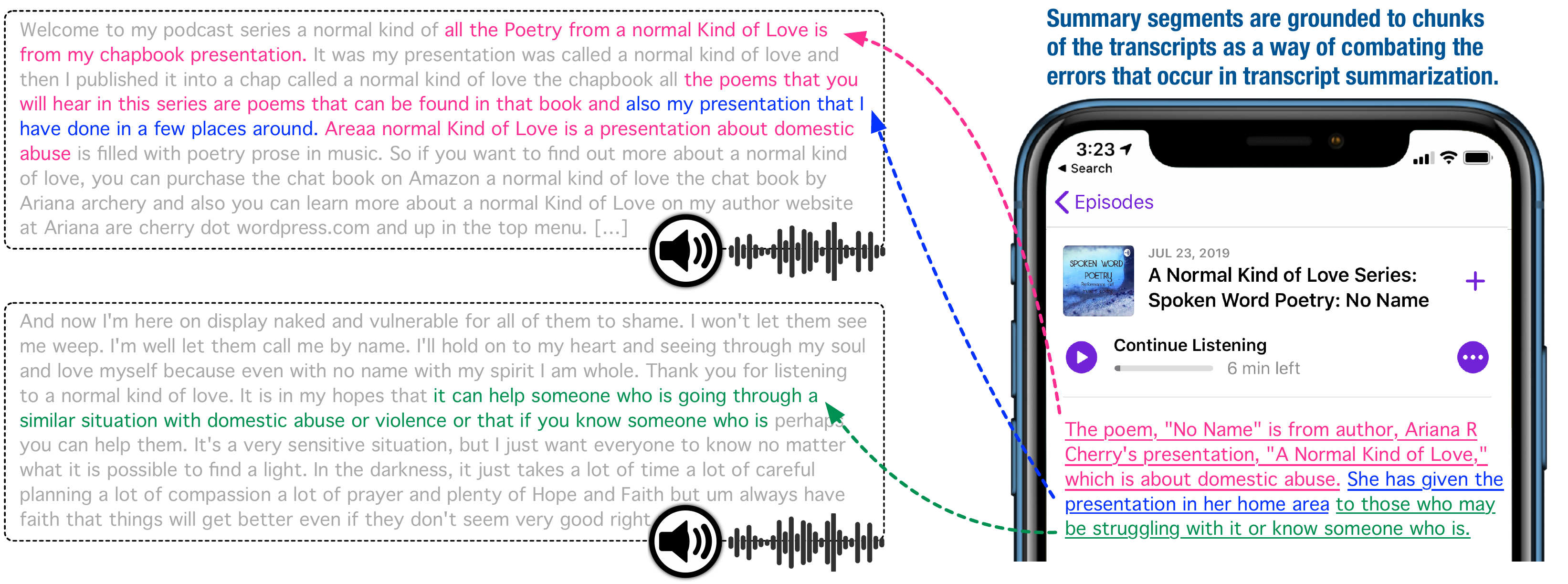}
\caption{An example of a \emph{grounded} summary where spans of summary text are tethered to the original audio.
The user can tap to hear the audio clip, thus interpreting a system-generated summary in context.
}
\label{fig:example}
\vspace{-0.1in}
\end{figure*}

It is desirable to generate \emph{grounded} summaries from podcast transcripts, where spans of summary text are closely tethered to the original audio.
Figure~\ref{fig:example} provides an example of a grounded abstractive summary.
When a user clicks on a summary segment, she will be directed to an audio clip that gives further detail of the conversational context. 
Grounded summaries give us a preview of notable podcast clips~\cite{Shalom:2019} and they may further release summarization service providers from potential legal claims by directing users to the original audio.
This is because, speech recognizers induce transcription errors and abstractive summarization models may hallucinate facts that are not entailed by the original~\cite{kryscinski-etal-2020-evaluating}, 
both can cause podcast summaries to contain misleading or inaccurate information. 
With grounded summaries, users are able to frame, interpret, and place into context any system-generated summaries, thus reducing the barriers to deploy podcast summarization technology.

One may attempt to align summary text and podcast transcripts in a post-processing step to generate grounded summaries.
Unfortunately, hallucinations do not allow for proper alignments as they are not found in the transcripts~\cite{maynez-etal-2020-faithfulness}.
Hierarchical attention models may seem promising for this task~\cite{liu-lapata-2019-hierarchical}.
However, the excessive length of the transcripts makes it difficult to produce attention distributions over the entire transcripts. 
Recent evidence suggests that attention weights are not reliable indicators of the relative importance of inputs~\cite{jain-wallace-2019-attention},
thus it remains an open question whether attention can be used to find alignments between transcripts and summary segments.

In this paper, we seek to generate grounded summaries from podcast transcripts by exploring an \emph{on-demand} abstractive summarizer.
It mimics how a human might approach a lengthy transcript -- the expert would identify a portion of the transcript that is deemed most important and relevant to the existing summary, 
use it as a ground to produce a new piece of the summary, and that process is repeated until the summary is finished.
Our summarizer employs a novel regularization technique that enables it to visit portions of the transcript in chronological order,
while allowing zigzags in order to produce a coherent summary.
This has another implication. 
It implies that we may estimate what percentage of a podcast transcript is covered by the summary and thus adjust that when necessary.

Distinguishing our work from earlier research on extract-then-abstract methods~\cite{hsu-etal-2018-unified,chen-bansal-2018-fast,gehrmann-etal-2018-bottom,lebanoff-etal-2019-scoring,jin-etal-2020-multi,pilault-etal-2020-extractive}, 
we require selected transcript chunks to have high salience, but also those salient content must appear at the beginning of the selected chunks,
so that the corresponding audio clips can provide good \emph{jump-in points} for users to start listening.
Our experiments are performed on a large podcast summarization dataset containing over 100,000 English podcasts~\cite{clifton-etal-2020-100000}.
We show that our proposed grounded summarizer can perform competitively or better than the state-of-the-art methods, 
including the recent methods that leverage large, pretrained models~\cite{lewis-etal-2020-bart,Beltagy2020Longformer} as judged by automatic metrics and human evaluation.
Our contributions in this paper are as follows.
\begin{itemize}[topsep=3pt,itemsep=-1pt,leftmargin=*]

\item We address the problem of podcast summarization by investigating an on-demand summarizer that produces grounded abstracts.
The abstracts help users quickly decide if they will listen to the podcasts and offer a sampler of salient podcast clips.
The on-demand summarizer does not need to encode the entire transcript, hence substantially reduces the GPU memory footprint. 

\item We conduct a series of analyses to gain insights into the impact of specific design decisions.
They include how a transcript chunk should be defined, 
whether those transcript chunks overlap, 
to what extent the summary content is taken verbatim from selected chunks, 
and how the summary may be extended to cover more information.

\item Through extensive experiments on a benchmark podcast dataset, we demonstrate the effectiveness of our proposed approach and show results that are comparable to human writer performance.  
The approach opens an avenue towards generating a new kind of abstractive summaries that allow users to verify the information consistency of summary parts against the original audio clips.\footnote{Our model and code have been made publicly available: \url{https://github.com/tencent-ailab/GrndPodcastSum}}

\end{itemize}

\section{Related Work}
\label{sec:related}

With the rapid rise of podcasts comes the need for automatic summarization of podcast transcriptions.
While comparatively understudied, recent work has shown great progress.
Clifton et al.~\shortcite{clifton-etal-2020-100000} present the Spotify dataset that was adopted in TREC 2020 for the podcast summarization task.\footnote{\url{https://trec.nist.gov/data/podcast2020.html}}
Our participating system in TREC 2020 focuses on identifying salient segments from transcripts and using them as input to an abstractive summarizer~\cite{song2020}.
Reddy et al.~\shortcite{reddy-etal-2021-detecting} develop classifiers to detect and eliminate extraneous marketing materials in podcasts to aid summarization.
In this paper, we explore techniques that generate \emph{grounded podcast summaries} where pieces of summary text are tied to short podcast clips.

One of the most serious problems of neural abstractive summarization is that the summaries can contain factually incorrect information and hallucinations~\cite{falke-etal-2019-ranking,kryscinski-etal-2020-evaluating,maynez-etal-2020-faithfulness,lebanoff-etal-2020-understanding}.
Without grounded summarization, users have to listen to the full episodes to find connections between details of the summaries and the original podcasts.
If successful, grounded summaries will benefit a number of summarization tasks where the input involves lengthy transcripts,
including meetings~\cite{li-etal-2019-keep,koay-etal-2020-domain,koay-etal-2021-sliding,zhong2021}, medical conversations~\cite{liu-chen-2019-reading}, interviews~\cite{zhu2021mediasum}, livestreams~\cite{cho-etal-2021-streamhover} and more.

An extract-then-abstract strategy could be used to produce grounded abstractive summaries~\cite{chen-bansal-2018-fast,gehrmann-etal-2018-bottom,hsu-etal-2018-unified,jin-etal-2020-multi,pilault-etal-2020-extractive}. 
Most of these approaches are tailored to written documents, e.g., news, Wikipedia, and scholarly articles.
They extract sentences from the documents and use them as input to an abstractive summarization model to produce a summary.
Nevertheless, transcripts of spoken language lack essential document structure such as sentence, paragraph and section boundaries, 
making it unclear how these approaches will perform on podcasts.

Attention provides another mechanism for aligning the summary and transcript segments. 
The use of sparse attention allows a summarization model to potentially scale to longer documents~\cite{Beltagy2020Longformer,Kitaev2020Reformer,huang2021efficient}.
Hierarchical Transformer encodes multiple paragraphs in a hierarchical manner to allow them to exchange information~\cite{liu-lapata-2019-hierarchical,fabbri-etal-2019-multi,chen-yang-2020-multi}.
However, it is shown that attention weights are not reliable indicators of the relative importance of inputs,
as alternative attention distributions would have yielded similar results~\cite{jain-wallace-2019-attention}.

Our approach in this paper is to better align summary segments with chunks of the transcripts to allow easy tracing of inconsistent information.
It features a generator that writes a summary from beginning to end, 
and a savvy selector that knows when to switch to a new transcript chunk and where to switch to.
Differing from PG networks~\cite{see-etal-2017-get} and retrieval-augmented generation~\cite{guu2020realm,lewis2021retrievalaugmented}, 
our selector places heavy emphasis on modeling and selection of transcript chunks.
A desirable chunk is expected to be about 2 minutes long and places important information at the beginning to enable easy user verification.
In the following section, we present details of the model implementation.

\section{Our Approach}
\label{sec:approach}

A major challenge facing podcast summarization is the dramatic length difference between source and target sequences.
At a speaking rate of 122 words per minute for spontaneous speech~\cite{polifroni-etal-1991-collection},
the full transcript of a 1-hour long episode contains roughly 7,000 words and that of a 1.5-hour long episode could reach 10,000 words.
In contrast, a podcast summary is short, containing on average 61 words according to Manakul and Gales~\shortcite{manakul2020}.
The ratio of their lengths could reach as high as 100-to-1, 
and this motivates our study of abstractive \emph{grounded} summarization
where summary segments are grounded to selected chunks of transcripts 
as a way of combating the inevitable errors that occur in podcast summarization.

Let $\mathbf{x}$ be the sequence of tokens in the source transcript and $\mathbf{y}$ be the sequence of tokens in the summary.
These tokens share the same vocabulary $\mathcal{V}$.
We use $\mathbf{x}_{\mathcal{C}}$ to denote a chunk of the transcript, and $\mathcal{C}$ gives the indices of tokens that belong to the chunk.
The full transcript can be decomposed into a sequence of chunks, denoted by $\{\mathcal{C}_1, \cdots, \mathcal{C}_M\}$.
The chunks may have varying sizes and overlap with each other; they are the grounds for generating a podcast summary.
Our assumption is twofold.
Firstly, we assume a summary segment is produced by conditioning on the previously generated tokens ($\mathbf{y}_{<j}$) and a specific chunk of the transcript.
Secondly, there exists a function $\mathcal{G}(\mathbf{x},\mathbf{y}_{<j})$ (Eq.~(\ref{eq:p_y_x})) that determines the most appropriate grounding chunk for generating all tokens of the segment.
Particularly, when the entire transcript is treated as a single chunk, it reduces to the standard conditional generation model $p_\theta(y_j | \mathbf{y}_{<j}, \mathbf{x})$.
\begin{align*}
p(\mathbf{y}|\mathbf{x}) = \prod_{j=1}^N p_\theta(y_j | \mathbf{y}_{<j},\mathcal{G}(\mathbf{x},\mathbf{y}_{<j}))
\numberthis\label{eq:p_y_x}
\end{align*}

Thus, the crucial point is a coarse segmentation of the source transcript and an alignment between the transcript chunks and summary segments.
In this work we use a sliding window to produce transcript chunks, with window size $\mathcal{W}$ and stride size $\mathcal{S}$.\footnote{
Discourse segmentation is beyond the scope of this work.
There is little to no data available to build a discourse segmentation tool and little existing work on discourse analysis of podcasts.
We refer the reader to Joty et al.~\shortcite{joty-etal-2019-discourse} for recent advances in discourse processing research.
}
The sizes can be measured in terms of tokens. 
E.g., $\mathcal{W}$=256 and $\mathcal{S}$=128 tokens will produce a series of fixed-length chunks that overlap with each other.
The rationale for using overlapping chunks is to find those that serve both as grounds for summary generation and good jump-in points for user verification. 
The sizes can also be measured by the number of sentences.
E.g., $\mathcal{W}$=20 and $\mathcal{S}$=20 sentences produce a set of varying-length, non-overlapping chunks.
In spoken language, a series of consecutive short sentences often indicates the content is relatively unimportant~\cite{marge-etal-2010-using}.

Given a summary segment $\widetilde{\mathbf{y}}$, we designate $\mathbf{x}_{\mathcal{C}}$ as a \textbf{grounding chunk} if it attains the highest score $\mathcal{S}(\mathbf{x}_{\mathcal{C}}, \widetilde{\mathbf{y}})$ (Eq.~(\ref{eq:coverage})).
This position-biased coverage score favors the transcript chunk that covers summary bigrams and puts summary content at the beginning to aid humans in performing content verification.
It measures the percentage of unique summary bigrams $\mathcal{B}(\widetilde{\mathbf{y}})$ covered by a chunk $\mathbf{x}_{\mathcal{C}}$.
Particularly, $\mathbbm{I}\left[ b_k \in \mathbf{x}_{\mathcal{C}} \right]$ is an indicator that returns 1 if the bigram $b_k$ appears in $\mathbf{x}_{\mathcal{C}}$ and 0 otherwise.
Each bigram $b_k$ has an associated weight $w_k$ (Eq.~(\ref{eq:weight})). 
If it appears in the first position of $\mathbf{x}_{\mathcal{C}}$ ($\texttt{pos}_k=0$), it receives a weight of one.
Otherwise, the weight is decayed according to the relative position of the bigram's first occurrence in the chunk ($\texttt{pos}_k$) and $\gamma$ is a coefficient for the decay.\footnote{
If a summary segment cannot be mapped to a chunk using Eqs.~(\ref{eq:coverage}-\ref{eq:weight}),
we perform the following:
$\widetilde{\mathbf{y}}$ is assigned to the first chunk $\mathcal{C}_1$ if it is the first segment of the summary.
Otherwise, $\widetilde{\mathbf{y}}$ is assigned to the same chunk as the previous summary segment to improve coherence.
We require $\mathbf{x}_{\mathcal{C}}$ and $\widetilde{\mathbf{y}}$ to have a minimum of four shared bigrams (stopwords-only bigrams are excluded).
Future work may consider aligning transcripts and summaries based on propositions~\cite{ernst2020superpal}.
}
\begin{align*}
& \mathcal{S}(\mathbf{x}_{\mathcal{C}}, \widetilde{\mathbf{y}}) = \frac{1}{|\mathcal{B}(\widetilde{\mathbf{y}})|} \sum_{b_k \in \mathcal{B}(\widetilde{\mathbf{y}})} w_k\, \mathbbm{I}\left[ b_k \in \mathbf{x}_{\mathcal{C}} \right] \numberthis\label{eq:coverage}\\
& w_k = 1 - \gamma\frac{\texttt{pos}_k}{|\mathcal{C}|}; \quad \gamma \in [0,1]  \numberthis\label{eq:weight}
\end{align*}

\begin{figure}
\centering
\includegraphics[width=3in]{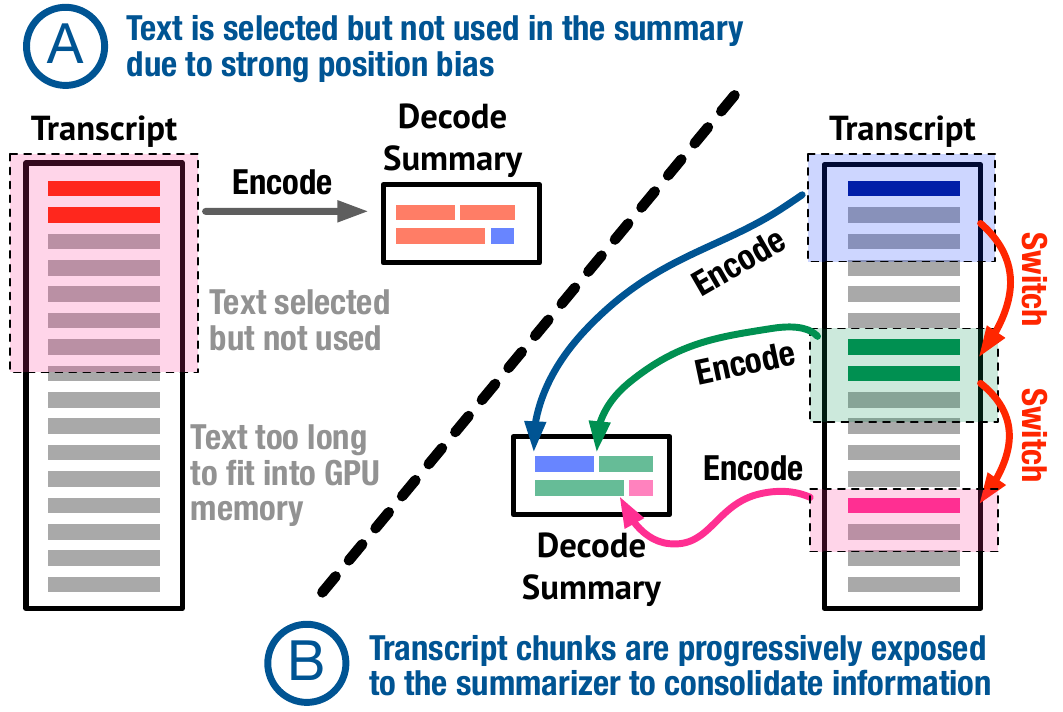}
\caption{Strong position bias can cause the abstractor to use only content at the beginning of the input to generate a summary.
By exposing the chunks progressively, our approach makes use of this characteristic to consolidate information from multiple transcript chunks.
}
\label{fig:architecture}
\vspace{-0.1in}
\end{figure}

We proceed by training a neural encoder-decoder model to generate an abstractive summary from the grounding transcript chunks.
Each segment of the summary (= sentence)\footnote{We use sentences as summary segments; other sentence-like segments are possible in future work.} is generated conditioned on its grounding chunk $\mathbf{x}_{\mathcal{C}}$ and all the previously generated tokens $\mathbf{y}_{<j}$.
The process starts from the first chunk of the transcript $\mathbf{x}_{\mathcal{C}_1}$.
The encoder converts this grounding chunk into a sequence of hidden vectors $[\mathbf{h}_1^{\mathcal{C}}, \ldots, \mathbf{h}_m^{\mathcal{C}}]$ (Eq.~(\ref{eq:encode})).
The decoder predicts the next summary token $y_j$ (Eq.~(\ref{eq:decode})) and continues to do so until a ``\emph{switch point}'' is detected.
At this point the current summary segment is finished and the decoder is poised to select the next transcript chunk $\mathbf{x}_{\mathcal{C}_{\texttt{new}}}$ and generate a new summary segment from it.
The decoding process finishes when a special symbol (\textsf{\footnotesize[sep]}) is predicted that indicates the end of the summary.
\begin{align*}
& [\mathbf{h}_1^{\mathcal{C}}, \ldots, \mathbf{h}_m^{\mathcal{C}}] = \texttt{Encode}(\mathbf{x}_{\mathcal{C}})\numberthis\label{eq:encode}\\
& y_j = \texttt{Decode}(\mathbf{y}_{<j}, [\mathbf{h}_1^{\mathcal{C}}, \ldots, \mathbf{h}_m^{\mathcal{C}}])\numberthis\label{eq:decode}\\
& \mathcal{G}(\mathbf{x},\mathbf{y}_{<j}) = 
\begin{cases}
\mathbf{x}_{\mathcal{C}_1}, & j = 1 \\
\mathbf{x}_{\mathcal{C}_{\texttt{new}}}, & j > 1\texttt{ \& }\texttt{switch} \\
\mathcal{G}(\mathbf{y}_{<j-1}), & j > 1\texttt{ \& }\texttt{no-switch}
\end{cases}
\end{align*}

There is a notable difference between our approach and most extract-then-abstract approaches that select important sentences from the document and provide them to the abstractor all-at-once.
As illustrated in Figure~\ref{fig:architecture}, 
strong position bias causes the abstractor to use only content at the beginning of the input to generate a summary.
By exposing the chunks progressively, our approach naturally makes use of this characteristic to consolidate information from multiple source chunks.
It reduces the amount of computation necessary to train the encoder-decoder model, as only selected transcript chunks are encoded which is equal to the number of summary segments.
Moreover, it is possible to encourage the summary to have a good coverage of the source content by specifying a minimal set of grounding chunks to be used for generation.

\vspace{0.08in}
\noindent\textbf{Regularizing Chunk Selection.}\quad
Learning function $\mathcal{G}(\mathbf{x},\mathbf{y}_{<j})$ that predicts a transcript chunk $\mathbf{x}_{\mathcal{C}}$ to switch to is crucial for success at inference time.
Let there be $M$ transcript chunks and $N$ summary segments in a training instance.
We define $p_j^c$ to be the model probability that the $c$-th chunk is predicted as the ground for generating the $j$-th summary segment;
$c^*$ is the gold chunk obtained using Eq.~(\ref{eq:coverage}-\ref{eq:weight}).
Our learning objective is a cross-entropy loss against the gold labels with a novel regularizing term $\mathcal{R}$ to enable chunks to be selected as per their original order in the transcript, while allowing zigzags to produce a coherent summary (Eq.~(\ref{eq:ret_loss}-\ref{eq:reg})).
\begin{align*}
&\mathcal{L}(\phi) = - \textstyle\sum_{j=1}^N \log p_j^{c^*} + \alpha \mathcal{R}\numberthis\label{eq:ret_loss}\\
&\mathcal{R} = \textstyle\frac{1}{N} \sum_{j=1}^N \sum_{c=1}^M \max(0, s_{j+1}^c - s_j^c)\numberthis\label{eq:reg}
\end{align*}

Particularly, $s_j^c = \sum_{c'=1}^c p_j^{c'}$ denotes the sum of the probability assigned to all chunks up to the $c$-th position, in order to generate the $j$-th summary segment. 
We encourage $\sum_{c=1}^M \max(0, s_{j+1}^c - s_j^c)$ to be a small value so that if a chunk (up to the $c$-th position) is assigned to the $j$-th summary segment, it is unlikely to be assigned to the ($j+1$)-th segment.
$\mathcal{R}$ is designed to regularize the loss and penalize violations;
$\alpha$ is its coefficient which will be tuned on the validation set.

Given a partial summary $\mathbf{y}_{<j}$, selecting the next transcript chunk depends on two factors.
Firstly, it should be a chunk that contains salient content at its beginning.
We use $\mathcal{I}\,(\mathbf{x}_{\mathcal{C}})$ to denote the importance of the chunk.
It is obtained by encoding the chunk into a vector $\mathbf{h}_{\mathbf{x}_\mathcal{C}}$ using RoBERTa~\cite{liu2019roberta}, then apply a feedforward network to it to estimate the importance (Eq.~(\ref{eq:imp})).\footnote{
The parameters of $\texttt{FFN}_1$ are pretrained on an extraction task that favors chunks that contain summary content at the beginning.
For each chunk, we compute its position-biased coverage score (Eq.~(\ref{eq:coverage})) against the entire summary.
1/4 of the chunks that yield the highest coverage scores are designated as positive instances, the remaining are negative instances.
$\texttt{FFN}_1$ is thus pretrained as a binary classifier.
}
\begin{align*}
p_j^{c} \propto & \exp(\mathcal{I}\,(\mathbf{x}_{\mathcal{C}}) + \mathcal{R}\,(\mathbf{x}_{\mathcal{C}},\mathbf{y}_{<j}))\numberthis\label{eq:ret}\\
\mathcal{I}\,(\mathbf{x}_{\mathcal{C}}) = &\texttt{ FFN}_1\,(\mathbf{h}_{\mathbf{x}_\mathcal{C}})\numberthis\label{eq:imp}\\
\mathcal{R}\,(\mathbf{x}_{\mathcal{C}}, \mathbf{y}_{<j}) = &\texttt{ FFN}_2\,([ \mathbf{h}_{\mathbf{x}_\mathcal{C}} || \mathbf{h}_{\mathbf{y}_{<j}} ])\numberthis\label{eq:rel}\\
+ &\texttt{ LowRank}\,(\mathbf{h}_{\mathbf{x}_\mathcal{C}}^\top \mathbf{W}\, \mathbf{h}_{\mathbf{y}_{<j}})
\end{align*}

Secondly, the chunk may be relevant to the partial summary $\mathbf{y}_{<j}$.
We define the relevance score $\mathcal{R}\,(\mathbf{x}_{\mathcal{C}}, \mathbf{y}_{<j})$ to capture two levels of interaction between the candidate chunk, represented by $\mathbf{h}_{\mathbf{x}_\mathcal{C}}$ and the last hidden state of the partial summary, represented by $\mathbf{h}_{\mathbf{y}_{<j}}$.
Their linear interaction is captured by a feedforward network ($\texttt{FFN}_2$) and bilinear interaction is modelled by $\mathbf{h}_{\mathbf{x}_\mathcal{C}}^\top \mathbf{W}\, \mathbf{h}_{\mathbf{y}_{<j}}$ 
where a low-rank approximation is used: $\texttt{LowRank}\,(p^\top \mathbf{W} q) = (p^\top \mathbf{U}) (\mathbf{V}^\top q)$.
The score $p_j^{c}$ is the likelihood that the $c$-th chunk is assigned to the $j$-th summary segment considering saliency and content relevancy.

\vspace{0.08in}
\noindent\textbf{Switch Point.}
A skilled writer pauses after writing down a sentence.
We borrow that intuition to inform the construction of a switch-point predictor.
The model combines the last hidden state of the summary sequence $\mathbf{h}_{\mathbf{y}_{<j}}$ and the embedding of the anticipated token $\mathbf{E}(y_{j})$, and use a feedforward network $\texttt{FFN}_3$ to predict if the $j$-th decoding step corresponds to a ``\emph{switch point}'' (Eq.~(\ref{eq:switch})).
During training, the last token of each summary sentence is a ground-truth switch point.
At inference time, the model predicts a switch point if $p(\texttt{switch})$ exceeds a threshold, 
at which point we compute $p_j^{c}$ to decide the next chunk.
Note that the model may choose use the same transcript chunk after switching.
\begin{align*}
p(\texttt{switch}) = & \sigma(\texttt{FFN}_3\,([ \mathbf{h}_{\mathbf{y}_{<j}} || \mathbf{E}(y_{j}) ])) \numberthis\label{eq:switch}
\end{align*}

\begin{table*}
\setlength{\tabcolsep}{3pt}
\renewcommand{\arraystretch}{1.15}
\begin{minipage}[t]{0.45\hsize}
\centering
\begin{fontpbk}
\begin{scriptsize}
\begin{tabular}[t]{|p{3.3in}|}
\hline
\textcolor{blue}{\textbf{Creator Description}}\quad
Tune in as Natalie and Jessica debate physical vs. chemical exfoliation options, and see what our ultimate verdict is on the best type and specific products we love!\\
\hline
\hline
\textcolor{blue}{\textbf{hk\_uu\_podcast1}}\quad
In this episode, Jessica and Natalie go head-to-head in the Great Exfoliation Debate! They each advocate their own type of exfoliator and try each other's products to see if they're worth the price difference. They also do a wine pairing and talk about the pros and cons of each of the products they tried.\\
\hdashline
\textcolor{blue}{\textbf{UCF\_NLP2}}\quad
In this weeks episode, Jessica and Natalie go head-to-head in the great exfoliation debate. They each advocate for their own type of exfoliator, and then try each other's products for 10 minutes to see what they think. We also talk about the pros and cons of each type of product and recommend a wine to pair with this episode. Santa Julia Winemakers Reserve Mountain Blend .\\
\hdashline
\textcolor{blue}{\textbf{cued\_speechUniv2}}\quad
In this episode of the Great Exfoliation Debate, Jessica and Natalie talk about their favorite types of exfoliators and the pros and cons of each of their favorite products. We also do a wine pairing and talk about the benefits and drawbacks of different types of chemical and physical exfoliation products.\\
\hline
\hline
\textcolor{blue}{\textbf{GrndAbs-tn}}\quad
\textcolor{orange}{Natalie and Jessica are back with another episode of Skincare} \textcolor{orange}{Somali A's.} \textcolor{olive}{This week we're talking about what we like to call the ``Great} \textcolor{olive}{Exfoliation Debate.''} \textcolor{magenta}{We'll also be doing our wine pairing this week.} \textcolor{darkblue}{Santa} \textcolor{darkblue}{Julia Winemakers' Reserve Mountain Blend (2016)}\\
\hline
\end{tabular}
\end{scriptsize}
\end{fontpbk}
\end{minipage}
\hfill
\begin{minipage}[t]{0.47\hsize}
\centering
\begin{fontpbk}
\begin{scriptsize}
\begin{tabular}[t]{|p{2.6in}|}
\hline
\textcolor{blue}{\textbf{GrndAbs-to}}\quad
\textcolor{orange}{This week we are talking about what we like to} \textcolor{orange}{call the ``Great Exfoliation Debate.''} \textcolor{olive}{Because we've got two} \textcolor{olive}{different points of view and we are going to Duke it out mano} \textcolor{olive}{a mano this week.} \textcolor{magenta}{We will also of course do our wine pairing} \textcolor{magenta}{because we are your Somali A's and this week we're going} \textcolor{magenta}{with something a little bit more aggressive...a little bit bold.}\\[0.7em]
\hdashline
\textcolor{blue}{\textbf{GrndAbs-sn}}\quad
\textcolor{orange}{In this episode, Natalie and Jessica debate the} \textcolor{orange}{pros and cons of exfoliation.} \textcolor{olive}{Exfoliation is this step in your} \textcolor{olive}{skincare routine that is taking off all the dead skin cells on} \textcolor{olive}{your face.} \textcolor{magenta}{And the point of Exfoliating is to reveal brighter,} \textcolor{magenta}{healthier skin while reducing the size of your pores.} \textcolor{darkblue}{In this} \textcolor{darkblue}{week's episode, we'll be discussing the pros, cons, and what} \textcolor{darkblue}{we think is the best way to exfoliate your skin.}\\[0.7em]
\hdashline
\textcolor{blue}{\textbf{GrndAbs-so}}\quad
\textcolor{orange}{Natalie and Jessica are back to debate the mer-} \textcolor{orange}{its of exfoliation.} \textcolor{olive}{This week, they are going mano a mano} \textcolor{olive}{and will be debating the pros and cons of using exfoliating on} \textcolor{olive}{your face.} \textcolor{magenta}{We will also do our wine pairing because we are} \textcolor{magenta}{your Somali A's and this week we're going with something} \textcolor{magenta}{a little bit more aggressive.} \textcolor{darkblue}{We would like to recommend} \textcolor{darkblue}{Santa Julia Winemakers' Reserve Mountain Blend.} \textcolor{cyan}{That is a} \textcolor{cyan}{Malbec and Cab Franc blend from 2016.} \textcolor{violet}{It's just a bit of a} \textcolor{violet}{middle of the road wine but super super tasty}\\[0.7em]
\hline
\end{tabular}
\end{scriptsize}
\end{fontpbk}
\end{minipage}
\caption{
Grounded abstractive summaries (\textsf{\footnotesize GrndAbs-*}) demonstrate a high level of specificity compared to summaries without grounding.
The latter contains more generic content.
The segments of grounded summaries are tethered to specific transcripts chunks. 
If a listener finds the summary segment interesting, they can tap to hear the selected segment in context.
}
\label{tab:example_output}
\vspace{-0.1in}
\end{table*}

\section{Podcast Data}
\label{sec:data}

With over 100,000 podcast episodes, the Spotify dataset~\cite{clifton-etal-2020-100000} is one of the largest corpora available for podcast search and summarization.
It encompasses a wide range of topics: travel, business, sports, book reviews, mysteries, guided meditations, nutrition and weight loss, among others.
Each episode is accompanied by an audio file, an automatic transcript generated by Google's Speech-to-Text API,\footnote{\url{https://cloud.google.com/speech-to-text}} and metadata provided by the podcast creator. 
We do not use the audio data in this paper.
Our summarizer takes as input a transcript and uses the creator-provided episode description as the reference summary.

\vspace{0.05in}
\noindent\textbf{Data Filtering.}\quad
Episode descriptions provided by podcast creators show wide variations in quality.
When noisy descriptions are used as reference summaries, they can cause a summarizer to hallucinate content.  
We conduct aggressive filtering of the training data to remove low-quality creator descriptions so as to maintain a balance between the amount of training examples available and quality of those examples.
We clean up reference summaries on the token-, sentence- and summary-level.
Tokens that correspond to URLs, email addresses, @mentions, \#hashtags, and those excessively long tokens (>25 characters) are directly removed from the summaries.
Each sentence in the summary is given a salience score that is the sum of IDF scores of its words.
A low score (<10) indicates the sentence contains few informative words and it is thus removed from the summary.
Finally, if, after sentence removal, the reference summary is too short or cannot be properly aligned to transcript chunks (\S\ref{sec:approach}), the instance is removed from the dataset.\footnote{
A summary is required to contain a minimum of 10 BPE tokens and have >2 shared bigrams with all of its grounding chunks.
Only words whose IDF scores are greater than 1.2 are considered when computing sentence salience scores.
}
This process filters out a substantial amount of low-quality reference summaries, yielding 40,302 episodes in the training set.
The Spotify dataset has a standard test set of 1,027 episodes and 179 of them are set for human evaluation. 

\vspace{0.05in}
\noindent\textbf{Baselines.}\quad
Our baselines consist of three of the best performing systems in the TREC 2020 competition on podcast summarization.
These systems were judged the best performing by both automatic metrics and human evaluation performed by NIST assessors. 
All systems make use of the BART-large model~\cite{lewis-etal-2020-bart}.
The model is tuned first on a news summarization dataset, i.e., CNN/DM or XSum, then fine-tuned on the podcast dataset.
Due to the long length of the transcripts, Karlbom and Clifton~\shortcite{karlbom2020} describe a combined Longformer-BART model 
that replaces the BART attention layers with attentions of Longformer~\cite{Beltagy2020Longformer};
their system is named \textsf{\small hk\_uu\_podcast1}.
Song et al.~\shortcite{song2020} develop an extractive module to select segments from transcripts, then integrate the extractor with BART abstractor to generate summaries (\textsf{\small UCF\_NLP2}).
Their baseline (\textsf{\small UCF\_NLP1}) directly truncates the transcript to the first 1,024 tokens.
Manakul and Gales~\shortcite{manakul2020} develop a similar baseline (\textsf{\small cued\_speechUniv3}) using the first 1,024 tokens.
Further, they perform sentence filtering using a hierarchical attention model (\textsf{\small cued\_speechUniv1/2/4}) and ensembles of models from different data shuffles and checkpoints (\textsf{\small cued\_speechUniv1/2}).
In this paper, our system is called \textsf{\small GrndAbs} for generating grounded abstracts.
It has 4 options: \textsf{\small -to, -tn, -so, -sn}, indicating the sliding window is defined in terms of tokens (\textsf{\small -t}) or sentences (\textsf{\small -s}), overlapping (\textsf{\small -o}) or non-overlapping (\textsf{\small -n}).
We obtain outputs from these competitive baselines and our system to examine both the successes and failures of these attempts.

\begin{table*}
\setlength{\tabcolsep}{6pt}
\renewcommand{\arraystretch}{1.2}
\centering
\begin{fontpbk}
\begin{footnotesize}
\begin{tabular}{|l|ccc|cc|c|}
\hline
\textbf{Run ID} & \textbf{R-1(\%)} & \textbf{R-2(\%)} & \textbf{R-L(\%)} & 
\textbf{BertS(\%)} & \textbf{BLEURT} & \textbf{SummL} \\
\hline
\hline
cued\_speechUniv1 & 30.54 & 11.25 & 21.05 &
84.17 & -0.7434 & 58.16 \\
cued\_speechUniv2 & 30.52 & 11.36 & 21.16 &
84.20 & -0.7491 & 56.93 \\
cued\_speechUniv3 & 28.44 & 9.55 & 19.52 &
83.77 & -0.7897 & 55.58 \\
cued\_speechUniv4 & 29.00 & 10.42 & 19.95 & 
83.99 & -0.7781 & 51.75 \\
UCF\_NLP1 & 30.09 & 12.07 & 21.75 & 
84.16 & -0.7508 & 57.35\\
UCF\_NLP2 & 30.44 & 11.99 & 21.67 & 
84.14 & -0.7382 & 57.85 \\
hk\_uu\_podcast1 & 29.02 & 10.70 & 20.66 & 
84.21 & -0.7992 & 44.63 \\
\hline
\hline
\textbf{GrndAbs-so} & 25.42 & 7.95 & 16.93 & 
82.62 & -0.8164 & 80.44 \\
\textbf{GrndAbs-sn} & 25.58 & 8.27 & 16.99 & 
82.64 & -0.8220 & 78.80 \\
\textbf{GrndAbs-to} & 25.79 & 8.38 & 17.15 & 
82.67 & -0.8028 & 82.98 \\
\textbf{GrndAbs-tn} & 25.79 & 8.25 & 17.20 & 
82.71 & -0.8130 & 79.90 \\
\hline
\end{tabular}
\end{footnotesize}
\end{fontpbk}
\vspace{-0.05in}
\caption{Results on the standard test set containing 1,027 episodes. 
Our evaluation metrics include ROUGE variants (R-1, R-2 and R-L), BERTScore and BLEURT.
We report the length of the summary (SummL) measured in words.
}
\label{tab:results_rouge}
\vspace{-0.1in}
\end{table*}

\section{Results and Analysis}
\label{sec:results}

\vspace{0.05in}
\noindent\textbf{Experimental Settings.}\quad
Our encoder-decoder model uses BART-large as the base model before fine-tuning it on the podcast dataset.
We use the AdamW~\cite{loshchilov2017decoupled} optimizer, where the momentum parameters are set to 0.9 and 0.999.
The regularizing coefficient $\alpha$ is tuned on the validation set in the range of $\{0, 0.01, \underline{0.1}, 1\}$.
For summary decoding, we use beam search with a beam size $K$=4 and a length penalty $p$=2.
Our sliding window, measured in terms of tokens or sentences, only contain whole sentences.
We use the Byte-Pair Encoding (BPE) tokenizer with a vocabulary size $\mathcal{V}$=50,265.
For transcripts and reference summaries, we use the SpaCy tool to segment them into sentences (model \textsf{\small en\_core\_web\_lg 2.2.5}).

\begin{table}
\setlength{\tabcolsep}{2.6pt}
\renewcommand{\arraystretch}{1.2}
\centering
\begin{fontpbk}
\begin{footnotesize}
\begin{tabular}{|l|cc|c|cc|}
\hline
& \textbf{E}$\boldsymbol\uparrow$ & \textbf{G}$\boldsymbol\uparrow$ & \textbf{E+G}$\boldsymbol\uparrow$ & \textbf{Fair}$\boldsymbol\downarrow$ & \textbf{Bad}$\boldsymbol\downarrow$\\
\hline
cued\_speechUniv2 & 22.09 & \textbf{51.36} & 73.45 & 22.67 & \textbf{3.88}\\
UCF\_NLP2 & 22.29 & 46.71 & 69.00 & 20.93 & 10.08\\
hk\_uu\_podcast1 & 18.60 & 45.93 & 64.53 & 25.78 & 9.69\\
creator\_description & 13.95 & 42.05 & 46.00 & 30.43 & 13.57\\
\hdashline
\textbf{GrndAbs-tn} & \textcolor{red}{\textbf{25.19}} & 50.58 & \textcolor{red}{\textbf{75.77}} & \textbf{20.16} & 4.07\\
\hline
\end{tabular}
\end{footnotesize}
\end{fontpbk}
\vspace{-0.05in}
\caption{Human evaluation results.
25\% of grounded abstractive summaries are rated as \emph{Excellent} and 76\% receive a rating of either \emph{Excellent} (E) or \emph{Good} (G).
}
\label{tab:results_human}
\vspace{-0.2in}
\end{table}

\begin{table*}
\setlength{\tabcolsep}{3.6pt}
\renewcommand{\arraystretch}{1.15}
\centering
\begin{minipage}{\textwidth}
\begin{small}
\begin{tabular}{|c|cccccccc|}
\hline
& Q1: People & Q2: People & Q3: Main & Q4: Podcast & Q5: Title & Q6: Summ & Q7: Good & Q8: Start/End\\
System & Names & Add Info & Topics & Format & Context & Redund & English & Points\\
\hline
creator\_description & 60.08 & 50.19 & 80.81 & 59.61 & 57.00 & 16.28 & 88.76 & 60.16\\
hk\_uu\_podcast1 & 64.15 & 47.29 & 85.63 & 57.62 & 58.95 & \textbf{10.85} & 94.76 & 70.35\\
UCF\_NLP2& 67.38 & 51.55 & 87.02 & 63.57 &62.52 &  14.40 &\textbf{95.15} & 71.71\\
cued\_speechUniv2& 69.12 & 50.67 & 87.98 & 64.73 & 63.62 & 12.87 & 94.93 & \textbf{77.00}\\
GrndAbs-tn& \textbf{75.15} & \textbf{64.47} & \textbf{89.73} & \textbf{69.51} & \textbf{66.15} & 17.09 & 94.55 & 73.35\\
\hline
\end{tabular}
\end{small}
\caption{
Average scores per human judgment of 179 testing summaries on 8 Yes/No questions.
An assessor quickly skimmed the episode, and made judgments for each summary of the episode.
``creator\_description'' represents the episode description.
``cued\_speechUniv2,'' ``UCF\_NLP2'' and ``hk\_uu\_podcast'' are the top-3 teams in the Podcast Challenge.
Our system ``GrndAbs-tn'' learns to produce abstractive summary while grounding summary segments in specific portions of the transcript to allow for full inspection of summary details.
}
\label{tab:human_eval}
\end{minipage}
\vspace{-0.1in}
\end{table*}

\vspace{0.05in}
\noindent\textbf{Example Summaries.}\quad
In Table~\ref{tab:example_output}, we provide a direct comparison of system summaries.
This podcast is hosted by Natalie and Jessica who call themselves ``\emph{Skincare Sommeliers.}'' 
The episode is named ``\emph{The Great Exfoliation Debate}.''
We find that grounded abstractive summaries (\textsf{\footnotesize GrndAbs-*}) have a higher level of specificity compared to summaries without grounding.
Segments of grounded summaries are tied to specific transcripts chunks. 
If a listener finds a summary segment interesting, they can tap to hear the selected summary segment in context.
Our baselines are highly competitive. 
Their summaries tend to contain more generic content.
The description provided by podcast creators is relatively short and at times it does not directly summarize the episode.
There are clear benefits in automatic summarization of podcasts, which can reduce the cognitive load and the time it takes for podcast creators to write the summary.

\vspace{0.05in}
\noindent\textbf{Automatic Metrics.}\quad
In Table~\ref{tab:results_rouge},
we report results on the standard test set containing 1,027 podcast episodes. 
The metrics include ROUGE~\cite{lin-2004-rouge} variants that compare system summaries with creator descriptions based on n-gram overlap.
Further, we experiment with recently developed metrics: BertScore~\cite{Zhang2020BERTScore} and BLEURT~\cite{sellam-etal-2020-bleurt} that draw on deep neural representations to evaluate generated text.
Our approach does not outperform the baselines in ROUGE evaluation against creator descriptions.
However, the gap has been substantially reduced when more advanced metrics (BertScore and BLEURT) are considered.
There are two possible explanations.
First, grounded summaries are about 50\% longer than plain abstractive summaries. 
Their average length is about 80 words per summary, yielding low precision scores.
Second, the quality of creator descriptions can be poor.
Jones et al.~\shortcite{jones2020} report only 40\% of such descriptions are of Good or Excellent quality, 
indicating future work may consider creating high-quality ground-truth summaries.
Among the four variants of our approach, we observe that their difference is not prominent. 
The token-based, non-overlapping windows (-tn) variant outperforms others in terms of R-1 and R-L.
This system is used in subsequent experiments and analyses.

\begin{table}
\setlength{\tabcolsep}{3pt}
\renewcommand{\arraystretch}{1.15}
\centering
\begin{scriptsize}
\textsf{
\begin{tabular}{|l|l|}
\hline
\textbf{Q1} & Does the summary include \textbf{names of the main people} (hosts, \\
& guests, characters) involved or mentioned in the podcast? \\[0.2em]
\hdashline
\textbf{Q2} & Does the summary give any \textbf{additional information} about \\
& the people mentioned (such as their job titles, biographies, \\
& personal background, etc)? \\[0.2em]
\hdashline
\textbf{Q3} & Does the summary include the \textbf{main topic(s)} of the podcast? \\[0.2em]
\hdashline
\textbf{Q4} & Does the summary tell you anything about \textbf{the format of} \\
& \textbf{the podcast}; e.g. whether it's an interview, whether it's a chat \\
& between friends, a monologue, etc? \\[0.2em]
\hdashline
\textbf{Q5} & Does the summary give you \textbf{more context on the title} \\
& of the podcast?\\[0.2em]
\hdashline
\textbf{Q6} & Does the summary contain \textbf{redundant information}? \\[0.2em]
\hdashline
\textbf{Q7} & Is the summary written in \textbf{good English}? \\[0.2em]
\hdashline
\textbf{Q8} & Are the \textbf{start and end of the summary} good sentence and \\
& paragraph start and end points? \\[0.2em]
\hline
\end{tabular}
}
\end{scriptsize}
\vspace{-0.05in}
\caption{
There are 8 yes-or-no questions asked about the summary quality. 
An ideal summary should receive a ``yes'' (1) for all questions but Q6.
}
\label{tab:eval_questions}
\vspace{-0.1in}
\end{table}

\begin{table*}
\setlength{\tabcolsep}{4pt}
\renewcommand{\arraystretch}{1.15}
\centering
\begin{small}
\begin{tabular}{|l|l|}
\hline\Tstrut
\textbf{Excellent} & The summary accurately conveys all the most important attributes of the episode, which could include topical \\
& content, genre, and participants. In addition to giving an accurate representation of the content, it contains \\
& almost no redundant material which is not needed when deciding whether to listen. It is also \\
& coherent, comprehensible, and has no grammatical errors. \\
\hdashline
\textbf{Good} & The summary conveys most of the most important attributes and gives the reader a reasonable sense of what the \\
& episode contains with little redundant material which is not needed when deciding whether to listen. Occasional \\
& grammatical or coherence errors are acceptable.\\
\hdashline
\textbf{Fair} & The summary conveys some attributes of the content but gives the reader an imperfect or incomplete sense of \\
& what the episode contains. It may contain redundant material which is not needed when deciding whether to \\
& listen and may contain repetitions or broken sentences.\\
\hdashline
\textbf{Bad} & The summary does not convey any of the most important content items of the episode or gives the reader an \\
& incorrect or incomprehensible sense of what the episode contains. It may contain a large amount of redundant \\
& information that is not needed when deciding whether to listen to the episode.\\[0.3em]
\hline
\end{tabular}
\end{small}
\vspace{-0.05in}
\caption{Guidelines for human evaluation of podcast summaries provided by TREC.
}
\label{tab:eval_criteria}
\vspace{-0.1in}
\end{table*}

\vspace{0.05in}
\noindent\textbf{Human Evaluation.}\quad
It is imperative to perform human evaluation given that creator-provided descriptions are of poor quality and ground-truth summaries are nonexistent.
We follow the TREC guidelines to ask human evaluators to assign each summary to one of the four grades: \emph{Excellent}, \emph{Good}, \emph{Fair} and \emph{Poor}.
The excellent summary will accurately conveys the most important content of the episode (topical content, genre, and participants). 
It should contain almost no redundant material, be coherent, comprehensible, and has no grammatical errors~\cite{jones2020}.
We also asked the human evaluators to answer 8 yes/no questions regarding the quality of the summary as \cite{jones2020} suggested, those questions are shown in Table~\ref{tab:eval_questions}.
We conduct these experiments on the test set containing 179 podcast episodes as \cite{jones2020} did, where each summary is evaluated by five Master workers recruited on the mechanical turk.
As shown in Table~\ref{tab:results_human}, we find that humans prefer the lengthier grounded abstractive summaries, which substantially outperform all baselines.
25\% of grounded abstractive summaries are rated as \emph{Excellent} and 76\% of them receive a rating of either \emph{Excellent} or \emph{Good}.
Table~\ref{tab:human_eval} shows the results of the 8 questions. Comparing to previous best systems, our grounded abstractive summaries have a significant performance gain on retrieving important information including People Names(+6.03\%), People Additional Information(+12.92\%), Main Topics(+1.75\%), Podcast Format(+4.78\%) and Title related context(+2.47\%) with slight redundancy.

\vspace{0.05in}
\noindent\textbf{Chunk Selection and Switch Point Prediction.} 

\noindent 
We are curious to know how well our system  performs on predicting grounding chunks: $\mathcal{G}(\mathbf{x},\mathbf{y}_{<j})$.
In this study, we assume switch points are known and report results on the validation set.
Our decoder starts from the first transcript chunk and predicts the next chunk at each switch point.
We find that it achieves an accuracy of 86.02\% on identifying ground-truth chunks.
Next, we examine the performance of switch point prediction.
On the validation set, we observe that the predictor achieves 98.75\%, 84.95\% and 91.33\%, respectively, for precision, recall and F-score.
Moreover, each summary has an average of 3.67 switch points.
A majority of the time (92.42\%) the model decides to use the current chunk to continue to decode the next summary segment.
At a small percentage (7.58\%) the model decides to find to a new grounding chunk.
We find 1.24 unique grounding chunks per summary.
The statistics suggest that identifying grounding chunks is crucial for summary generation.

\vspace{0.05in}
\noindent\textbf{Grounded Summaries.}\quad
In Table~\ref{tab:results_ngram}, we measure the percentage of summary n-grams that appear in the transcripts (for all baselines) or grounding chunks (for our approach). 
While the distributions of unigrams are largely similar, we observe that grounded abstractive summaries tend to reuse more bigrams and trigrams of their grounding chunks.
Moreover, for trigrams that are found in the grounding chunks, 
we find 70\% of them tend to appear at the beginning -- the front half of the chunks.
These results suggest that the grounding chunks identified by our approach can provide effective support for summary generation.

\begin{table}
\setlength{\tabcolsep}{6pt}
\renewcommand{\arraystretch}{1.2}
\centering
\begin{fontpbk}
\begin{footnotesize}
\begin{tabular}{|l|ccc|}
\hline
& \textbf{1-gram} & \textbf{2-gram} & \textbf{3-gram}\\
\hline
\hline
cued\_speechUniv1 & 87.86 & 56.33 & 33.37 \\
cued\_speechUniv2 & 87.82 & 56.11 & 32.72 \\
cued\_speechUniv3 & 84.96 & 52.05 & 31.24 \\
cued\_speechUniv4 & 85.23 & 51.44 & 29.88 \\
UCF\_NLP1 & 83.96 & 49.89 & 28.61 \\
UCF\_NLP2 & 84.46 & 50.33 & 29.33 \\
hk\_uu\_podcast1 & 86.94 & 56.12 & 35.55 \\
\hline
\hline
\textbf{GrndAbs-to} & 85.83 & 57.31 & \textbf{39.03} \\
\textbf{GrndAbs-tn} & 86.38 & 58.44 & \textbf{40.38} \\
\textbf{GrndAbs-so} & 86.52 & 59.76 & \textbf{42.13} \\
\textbf{GrndAbs-sn} & 86.80 & 60.55 & \textbf{43.18} \\
\hline
\end{tabular}
\end{footnotesize}
\end{fontpbk}
\vspace{-0.05in}
\caption{Percentage of summary 1/2/3-grams appearing in the transcripts (for all baselines) or grounding chunks (for our approach). 
We observe that grounded abstractive summaries tend to reuse bigrams and trigrams of their grounding chunks.
}
\label{tab:results_ngram}
\vspace{-0.2in}
\end{table}

\vspace{0.05in}
\noindent\textbf{What Made the Task Challenging?} 

\noindent We manually analyze a large amount of transcripts and their creator descriptions to identify the challenging points of podcast summarization in Table~\ref{tab:example_reference}:
\begin{itemize}[topsep=3pt,itemsep=-1pt,leftmargin=*]
\item
Substantial lexical mismatch exists between the spoken and written form of descriptions. 
Speech recognition errors are abundant. 
E.g., ``\emph{by Hans Christian Andersen}'' has been misrecognized into ``\emph{buy homes Christian Andersen}.''

\item 
The creator descriptions are sometimes highly abstractive, do not always summarize the episode and contain teasers. E.g., ``\emph{A male perspective podcast to start a conversation...}'' 
and ``\emph{Ever wondered how Ed Sheeran became famous}.'' 

\item
The transcripts contain advertising inserts, e.g., ``\emph{I need to tell you about our sponsor...}''
and the same description is used for different episodes that causes confusion to the model, e.g., ``\emph{The goal of Daily Fortnite is to build a community...}'' 

\end{itemize}

\section{Conclusion}
\label{sec:conclusion}

In this paper, we investigate podcast summarization to produce textual summaries for podcast episodes that help listeners to understand why they might want to play those podcasts.
We present a new kind of podcast summary where spans of summary text are tethered to the original audio to allow users to interpret system-generated abstracts in context.
Experiments on a benchmark dataset demonstrates the utility of our proposed approach.

\section*{Acknowledgments}

The authors would like to thank all anonymous reviewers for their insightful comments which helped improve this paper.
This research was supported in part by the National Science Foundation (NSF) Grant \#2143792.

\bibliography{main}
\bibliographystyle{acl_natbib}

\appendix

\section{Appendix}
\label{sec:appendix}

\noindent \textbf{What made the task of podcast summarization challenging?}

\begin{table}
\setlength{\tabcolsep}{3pt}
\renewcommand{\arraystretch}{1.15}
\begin{minipage}[t]{0.45\hsize}
\centering
\begin{fontpbk}
\begin{scriptsize}
\begin{tabular}[t]{|lp{2.6in}|}
\hline
& \textcolor{darkblue}{\textbf{Lexical Mismatch between Written and Spoken Text}}\\
\hdashline
\textbf{[S]} & ASMR reading of \textcolor{red}{The Snow Man by Hans} \textbf{Christian Andersen}, 1861.\\
\textbf{[T]} & 
Hello, my darling. \emph{I need to tell you about our sponsor anchor dot f m. Anchor is a podcast creation and distribution tool. And it gives you everything you need to record edit. Plus they'll distribute your podcast to all of the major channels including Spotify Apple podcasts and Google podcasts free of charge you can make money with no minimum listenership and it couldn't be easier. Download the anchor app or go to Anchor dot f m-- to get started sweet dreams.} Hello, my darling and Welcome to our story time. For the 12 Days of Christmas. \textbf{Our next story is} \textcolor{red}{the Snowman buy homes} \textbf{Christian Andersen} and we have our warm and toasty fireplace to keep us cozy while I read to you if you like what you hear [...]\\ 
\hline
\hline
& \textbf{\textcolor{darkblue}{Highly Abstractive Reference Summary}}\\
\hdashline
\textbf{[S]} & \textcolor{red}{A male perspective podcast} \textbf{to start a conversation for men out there to begin the healing process} of what they bottle up inside.\\
\hdashline
\textbf{[T]} & 
[...] And you know what? I'm tired and I've sat down with a lot of guys in the past year a lot of women in the past year. I've shared my ideas with them and I really just want to inspire people \textbf{to start a conversation to help them begin the healing process} of you know, what I don't want to hold up things on the inside anymore. So I've been thinking about this word feelings feelings feelings feelings [...]\\ 
\hline
\hline
& \textbf{\textcolor{darkblue}{Same Summary for Different Podcast Episodes}}\\
\hdashline
\textbf{[S]} & The goal of Daily Fortnite is to build a positive community of Fortnite players so we can all enhance our enjoyment of Fortnite together.\\
\hdashline
\textbf{[T]} & Welcome back to another episode of daily Fortnight your daily podcast about Fortnight. I'm your host Mikey AKA Mike. Daddy AKA magnificant Mikey. So today we have the fishing frenzy results are in you can go check that out on the leaderboard [...] \\ 
\hline
\hline
& \textbf{\textcolor{darkblue}{Part of the Summary is Irrelevant to the Transcript}}\\
\hdashline
\textbf{[S]} & If you're like me \textbf{you sometimes suffer from ``imposter syndrome''.} 
\textcolor{red}{I hope these short positive messages will help my tile contractor friends to know their worth, overcome ``imposter syndrome'' and continue to grow their contracting businesses!}\\
\hdashline
\textbf{[T]} & 
[...] I will be doing a brief, you know podcast episode every week on mindset and I'm thinking of calling it mindset Monday [...]
\emph{The other thing I want to talk to you today about is a new sponsor of tile money. So I want to thank that my new sponsor} [...] 
So this new mindset segment that I want to record for the for the podcast episodes. You know, it got me thinking recently Chris Ford posted up. A question to the group about this thing called \textbf{impostor syndrome and it's something so many of us I struggle with it myself personally.}\\
\hline
\hline
& \textbf{\textcolor{darkblue}{Potential Teaser Texts in the Summary}}\\
\hdashline
\textbf{[S]} & \textcolor{red}{Ever wondered how Ed Sheeran became famous or how Stormzy writes his songs?} [...]
\textbf{Straight Up}, a game-changing \textbf{new podcast pulling back the curtain on UK music at its most exciting moment yet}, lifts the lid on all this and more.\\
\hdashline
\textbf{[T]} & 
\textbf{This is straight-up the 490 UK music podcast} hosted by journalists me cackling Johnston. I met Eleanor Halls will be taking you through the biggest music headlines the hottest entry closet and spotlighting the artists that we're into right now [...]
\textbf{our guests will pull back the curtain on the musicians that everyone's talking about to top it all off.} We chat all of our guests over their favorite drink. So why not grab a glass and join us for the stories? [...]\\ 
\hline
\end{tabular}
\end{scriptsize}
\end{fontpbk}
\end{minipage}
\caption{
What made the task of podcast summarization challenging?
a) Lexical mismatch between spoken and written forms and speech recognition errors (``\emph{by Hans Christian Andersen}'' was mistranscribed into ``\emph{buy homes Christian Andersen}.'')
b) Highly abstractive creator description, e.g., ``\emph{A male perspective podcast to start a conversation...}''
c) The same summary is used for different podcast episodes, e.g., ``\emph{The goal of Daily Fortnite is to build a positive community...}'' 
d) The creator description does not summarize or describe the episode, e.g., ``\emph{I hope these short positive messages will help my tile contractor friends...}'' and ``\emph{Ever wondered how Ed Sheeran became famous}''.
e) The podcast is improvised, its content lacks discourse structure, the transcript contains frequently recognition errors and advertising inserts, e.g., ``\emph{I need to tell you about our sponsor...}''
}
\label{tab:example_reference}
\end{table}

\end{document}